\documentclass[letterpaper]{article} 
\usepackage{aaai25}  
\usepackage{times}  
\usepackage{helvet}  
\usepackage{courier}  
\usepackage[hyphens]{url}  
\usepackage{graphicx} 
\usepackage{natbib}  
\usepackage{caption} 
\usepackage{algorithm}
\usepackage{algorithmic}
\usepackage{booktabs}
\usepackage{array}  
\usepackage{ragged2e}  
\usepackage{xcolor,colortbl}
\usepackage{graphicx} 
\usepackage{xcolor}

\usepackage{newfloat}
\usepackage{listings}
\DeclareCaptionStyle{ruled}{labelfont=normalfont,labelsep=colon,strut=off} 
\floatstyle{ruled}
\newfloat{listing}{tb}{lst}{}
\floatname{listing}{Listing}
\title{An Annotated Corpus of Arabic Tweets for Hate Speech Analysis}
\author{ 
    Wajdi Zaghouani\textsuperscript{\rm 1},
    Md. Rafiul Biswas\textsuperscript{\rm 2}
}
\affiliations{
    \textsuperscript{\rm 1}Northwestern University in Qatar, Education City, Doha, Qatar\\
    \textsuperscript{\rm 2}Hamad Bin Khalifa University, Doha, Qatar\\
}

\usepackage{bibentry}

\begin{document}

\maketitle

\begin{abstract}
Identifying hate speech content in the Arabic language is challenging due to the rich quality of dialectal variations. This study introduces a multilabel hate speech dataset in the Arabic language. We have collected 10000 Arabic tweets and annotated each tweet, whether it contains offensive content or not. If a text contains offensive content, we further classify it into different hate speech targets such as religion, gender, politics, ethnicity, origin, and others. A text can contain either single or multiple targets. Multiple annotators are involved in the data annotation task. We calculated the inter-annotator agreement, which was reported to be 0.86 for offensive content and 0.71 for multiple hate speech targets. Finally, we evaluated the data annotation task by employing a different transformers-based model in which AraBERTv2 outperformed with a micro-F1 score of 0.7865 and an accuracy of 0.786.
\end{abstract}

\section{Introduction}

Identifying hate speech has become a critical area of research due to the rise of online platforms where offensive and harmful content can spread rapidly. The engagement of communities in digital platforms is increasing day by day, and so it increases the risk of individuals being affected by the negativity of hate speech content \cite{costello2020hate}. Therefore, the development of an effective computational tool is required to monitor and mitigate such content. While significant advancements have been made in hate speech detection for English and other widely spoken languages, addressing this issue in the Arabic language remains particularly challenging due to the language's complexity and diversity \cite{alhazmi2024systematic}.

Arabic is a rich language with numerous dialectal variations across different regions, making natural language processing (NLP) tasks, including hate speech detection, more difficult than in languages with more standard forms \cite{alhazmi2024code}. Moreover, Arabic dialects are highly informal in nature, and they vary significantly from Modern Standard Arabic (MSA), adding layers of complexity when developing NLP models \cite{abdelsamie2024comprehensive,al2023detection}. Many of the available Arabic datasets focus on formal Arabic or a specific dialect, which limits the generalizability of hate speech detection models.

In addition to linguistic diversity, the low-resource nature of Arabic dialects also poses a significant challenge \cite{abdul2024survey}. Many Arabic NLP tasks suffer from a scarcity of annotated datasets, especially tasks like hate speech detection. While hate speech detection in English has benefited from large-scale datasets and well-developed models, similar efforts in Arabic are comparatively limited \cite{alkomah2022literature, kansok2023systematic}. The multilabel nature of hate speech—where a text can target multiple identities such as religion, gender, or ethnicity—adds further complexity, making it essential to develop datasets that capture the breadth of hate speech in different forms \cite{zaghouani2024so}.

This study addresses these challenges by introducing a novel multilabel hate speech dataset in Arabic, specifically designed to encompass multiple hate speech targets such as religion, gender, politics, ethnicity, and origin. By annotating a large corpus of Arabic tweets, we aim to provide a resource that can enable more accurate and nuanced detection of hate speech across different target categories. Our dataset is a significant contribution to the Arabic NLP field, providing an essential resource for future research on hate speech detection in low-resource and dialectally diverse languages. 

\textbf{1 Novelty:} The dataset represents a unique, large-scale, annotated corpus of Arabic tweets specifically designed for multilabel hate speech analysis. It addresses the paucity of resources in this domain and provides a foundational dataset for the training and evaluation of machine learning models.

\textbf{2 Availability:} The annotated dataset and guidelines are made freely accessible to the research community upon request. The dataset is released under the Creative Commons Attribution 4.0 International (CC BY 4.0) license, which allows for free use, distribution, and adaptation, provided that the original work is properly attributed. This open access ensures that the majority of academic and industry researchers can access and utilize the resource.

\textbf{3. Utility:} The dataset and guidelines are well-documented, with clear instructions for accessing and using the resource. The annotation process, which involves diverse annotators from various Arab countries, is described in detail, ensuring the representativeness and accuracy of the dataset. The corpus can be employed to train and evaluate various machine learning models of hate speech detection in Arabic social media content. \cite{kakulapati2023managing,xian2010framework,gaind2019emotion,pereira2019detecting}.

\section{Related Works}
Numerous research efforts have been conducted to develop and
examining Arabic corpora for various NLP applications. However, a few of them focused on the collection of hate speech datasets in Arabic. We will categorize the related works into two categories: i) collection of Arabic hate speech corpus and ii) techniques to detect Arabic hate speech. 

\textbf{Collection of Arabic Hate Speech Corpus}: 
Table \ref{tab:datsets-compare} describes a recent dataset on Arabic hate speech collections. \citealt{mulki-etal-2019-l} collected Arabic Twitter data between March 2018 and February 2019. They collected 5846 tweets containing three labels: normal, abusive, and hate. The search keywords contain refugee, female, Arabs, Druze, tweets belonging to the post, and replies from politicians, feminists, and activists. 
\cite{Albadi2018are} collected 9,316 tweets classified into three labels: hateful, abusive, or normal. The search keywords contain racist, religious, and ideological hate speech, and the search spans from March 2018 to August 2018. \cite{zaghouani2024so} collected tweets between between August 12, 2020, and October 4, 2020. It covers diverse contents focusing on emotion, emotion intensity, offensive content, fact-checing and so on. 
\cite{alshaalan2020hate} collected 3075 tweets in 2018 on religion, ethnicity, nationality, and gender. The tweet Id  belongs to gulf countries.  
\citealt{khezzar2023arhatedetector} proposed the arHateDataset dataset, which comprises 34,107 tweets obtained from different Arabic standards and dialectals. However, this dataset is not a unique corpus from the author. Rather, they combined seven Arabic datasets together to form this dataset. 

\begin{table*}[ht]
\centering
\begin{tabular}{>{\RaggedRight}p{2.5cm} >{\RaggedRight}p{2.5cm} c c c c>{\RaggedRight}p{2cm}}
\toprule
\textbf{Dataset} & \textbf{Keywords} & \textbf{Multilabel} & \textbf{Size} & \textbf{Platforms} & \textbf{Dialect} & \textbf{Model}\\ 
\midrule
\cite{mulki-etal-2019-l} & Refugee, female, Arabs & No & 5,846 & Twitter & Syrian,Lebanese & SVM, NB\\ 
\cite{Albadi2018are} & Racist, religious, tribes & No & 9,316 & Twitter & Saudi & SVM,CNN, RNN, BERT\\ 
\cite{zaghouani2024so} & Variety of topics & Yes & 15,965 & Twitter & Across Arab countries & LR, SVM, AraBERT\\ 
\cite{alsafari2020hate} & Religion, ethnicity, gender & No & 3,075 & Twitter & Gulf countries & LR, CNN, mBERT\\ 
\cite{khezzar2023arhatedetector} & Varieties & No & 34,107 & Twitter & Multi & NB, CNN, AraBERT\\ 
\bottomrule
\end{tabular}
\caption{Comparison of Arabic Hate Speech Datasets}
\label{tab:datsets-compare}
\end{table*}

\textbf{Techniques to Detect Arabic Hate Speech:}  Table \ref{tab:datsets-compare} demonstrates the uses of different model to detect offensive content. 
\cite{mulki-etal-2019-l} used traditional machine learning techiniques Support Vector Machine (SVM) and Naïve Bayes (NB) to identify tweets wheather a tweet is hateful or not. \cite{Albadi2018are} applied SVM and Logistic Regression (LR) techqniue as traditional machine learning model. For deep learning model they used CNN, RNN, GRU, CNN+GRU and for transformer, they applied BERT model. CNN outperformed on all of these experiments.  \cite{zaghouani2024so} applied different machine learning techniques such as LR, SVM, Random Forest (RF) and transformer based model AraBERT to classify the content either offensive or not. AraBERT scored for micro F1 score 0.65. \cite{alsafari2020hate} fused transforemr BERT base model with mahine learning model (LR and SVM) and with deep learning model (CNN, GRU, LSTM). The fusion model mBERT+CNN outperformed (micro F1.Score 0.7899) over other techniques. \cite{ khezzar2023arhatedetector} applied various techniques for hate speech classification and AraBERT obtained best accuracy score (0.91) among them.

These studies highlight the importance of developing annotated Arabic corpora for reliable hate speech detection within the challenges of informal dialectal social media posts. To extend the existing Arabic corpus, our dataset provides multilabel hate speech which is unique. Annotating each tweet to all the possible labels has been proposed for the first time in our dataset. Our work builds on prior literature through a large-scale annotated dataset and experiments with transformers like AraBERT. 

\section{Methodolgy}
\subsection{Data Collection}
\begin{figure}[h]
    \centering
    \includegraphics[width=\linewidth]{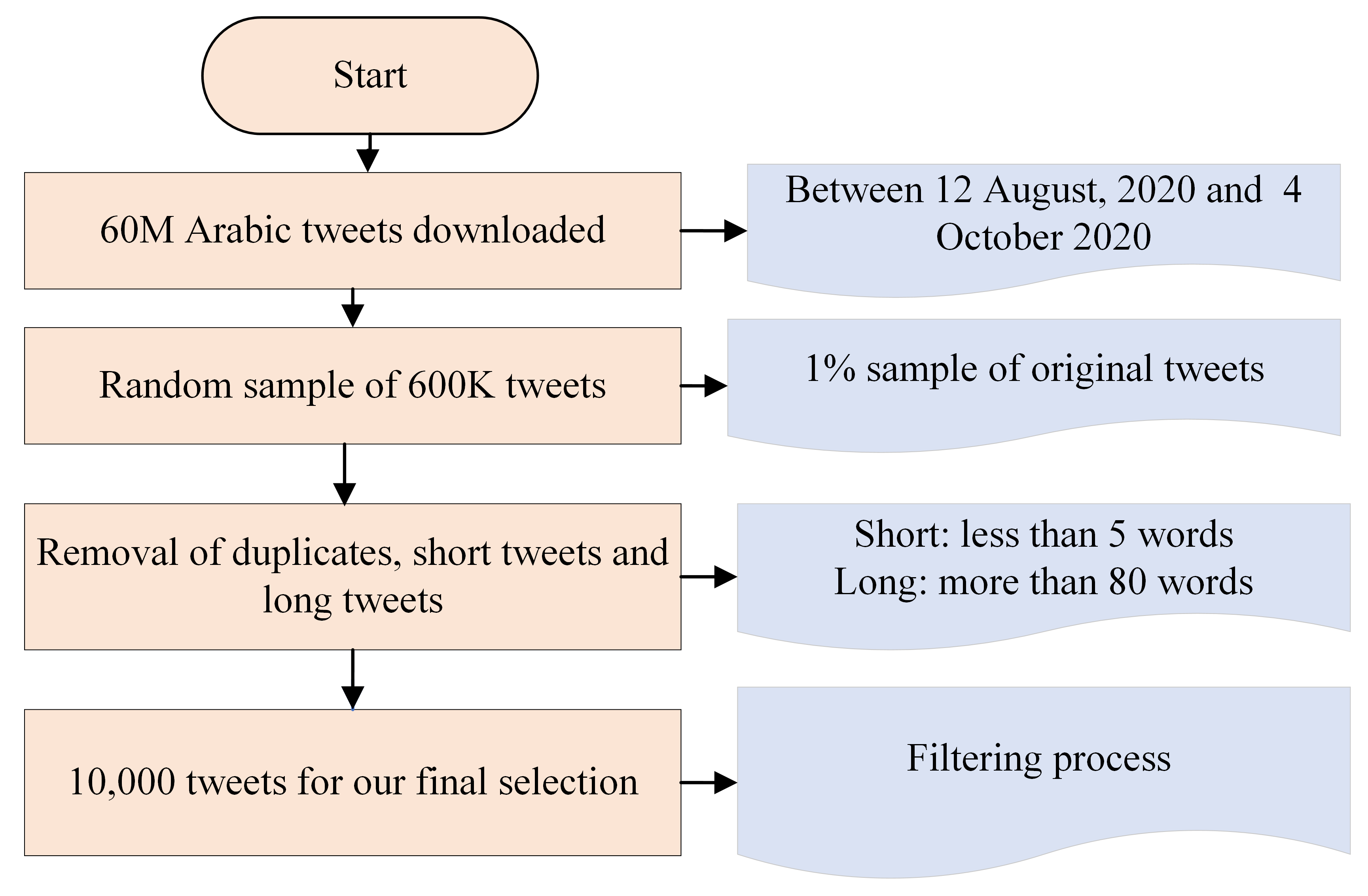}
    \caption{Data Collection}
    \label{fig:data-collection}
\end{figure}

We collected 60 million Arabic tweets from August 12, 2020, to October 4, 2020, and randomly sampled 600,000 tweets (1\% of the original). After removing duplicates, incomplete tweets, short tweets (less than 5 words), and long tweets (more than 80 words), we obtained a diverse set of tweets for annotation. In this current study, we used ASAD tools \cite{hassan2021asad} to classify offensive content from raw tweets. ASAD is a suite of tools for analyzing Arabic social media content, primarily focused on Twitter. It offers modules for tasks such as Dialect Identification, Sentiment and Emotion Analysis, Offensive Language and Hate Speech Detection, Spam Detection, and more. We identified offensive content by choosing 4000 tweets from the highest confidence (80\%-100\%), 4000 tweets from average confidence (60\%-79\%), and 2000 tweets from low confidence (1\%-39\%). Then, we applied sentiment classification and randomly selected 4000 tweets from positive, 4000 tweets from negative, and 2000 from neutral categories. Next, we selected tweets that were marked as hate speech by ASAD tools. We also included tweets that were labeled as adult content and spam. Figure \ref{fig:data-collection} shows the data collection process.

It is noteworthy that there exists a degree of overlap among the extracted tweets from the aforementioned selection steps. For instance, a tweet might simultaneously contain offensive content and adult content. We removed duplicates while merging the tweets and were left with 10,000 tweets. For this study, we focused on emotions and multilevel classification of offensive/hate speech content.

\subsection{Ethical Approval}
The authors carefully considered the ethical implications of the resource and applied for and received an institutional IRB exemption. The dataset was collected in compliance with Twitter (X) API rules, and user privacy was protected through the anonymization of personally identifiable information. The study adheres to established ethical guidelines for social media research.

\subsection{Annotation Procedure}
The annotation task was performed by several annotators (ranging from one to five based on the understanding of the dialects) and a manager (who led the project). 

\textbf{Annotators Recruiting and Training}: The annotators were recruited from different Arab countries, including two from Tunisia, one from Egypt, one from Jordan, and one from Qatar. Among them, three were female, and two were male. This diverse representation ensures that the dataset accurately captures the linguistic diversity and cultural contexts within the Arab-speaking world. Annotators from different countries provide insight into how certain words or phrases might be interpreted differently across cultures, improving the quality and accuracy of annotations.
The annotators underwent a training process to ensure high-quality and consistent annotations. Each category had practice questions to familiarize them with the types of content they would encounter. Annotators met regularly to discuss difficulties and revise their approaches as needed.

\textbf{Role of the Annotation Manager}: The annotation manager evaluated the annotation process. He resolved if there were any conflicts among annotators and chose the final label. The manager also analyzed errors, calculated inter-annotator agreement, provided feedback, and updated the guidelines to maximize the consistent label.

\textbf{Annotation Platform}: Annotation was performed using MicroMappers, an online tool to manage annotation. Annotators worked individually using this tool but met weekly to discuss their outcomes and difficulties.

\subsection{Annotation Guidelines}
Proper annotation guidelines are essential to building an appropriate dataset with fewer errors \cite{zaghouani2014large}. Also, a proper guideline can minimize subjective bias among annotators \cite{rottger2021two}. Figure \ref{fig:Corpus-Description} depicts a glimpse of annotation guidelines. This annotation is a multilabel procedure.

\begin{figure}[h]
    \centering
    \includegraphics[width=\linewidth]{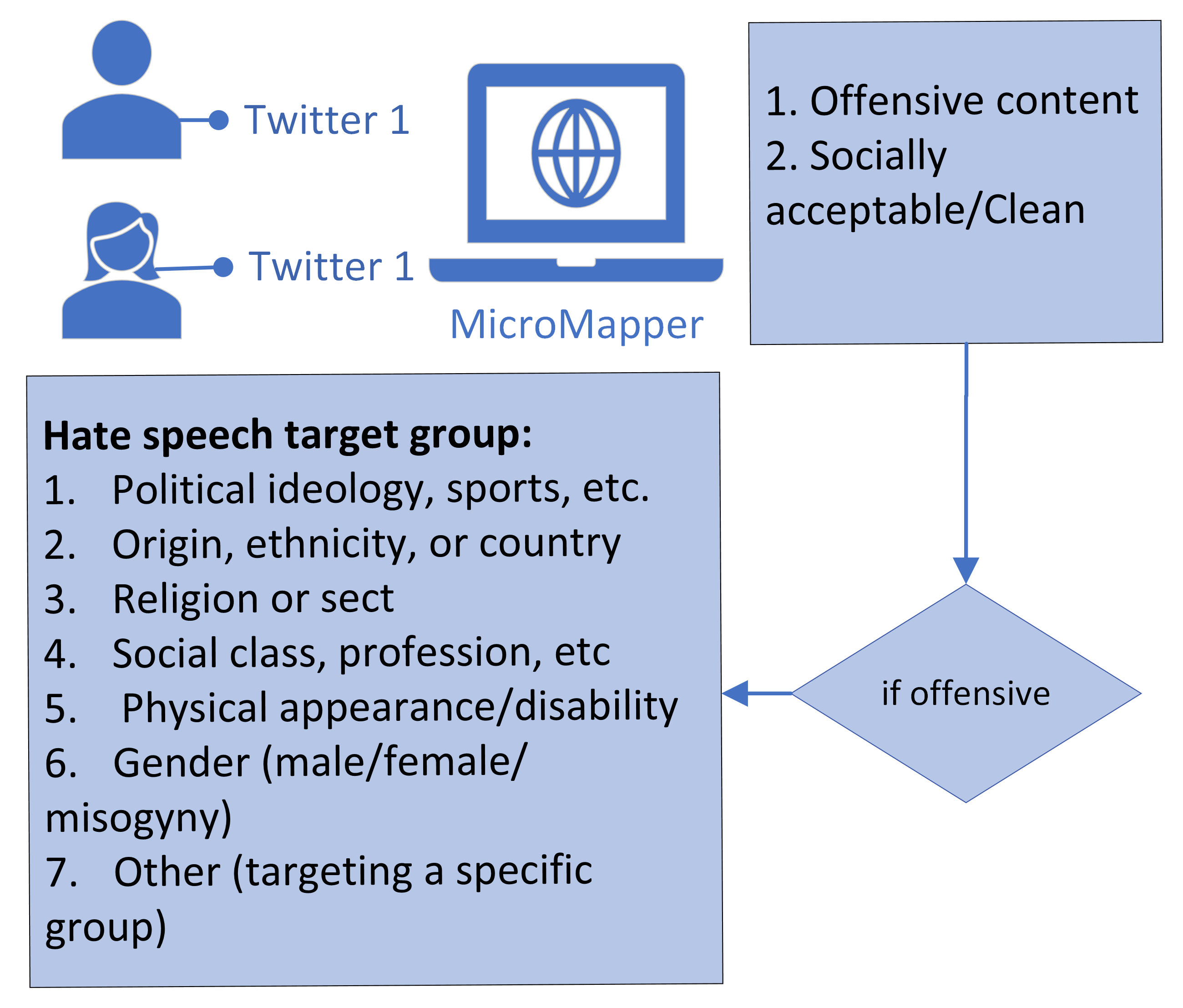}
    \caption{Data Collection}
    \label{fig:data-annotation}
\end{figure}

\textbf{Offensive:} Offensive content refers to cause harm, insult, or discomfort to individuals or groups. \\
Example: "This is a lie from you and has no source except from a seller among you".

\textbf{Socially acceptable\textbackslash Clean:} Tweet content does not carry offensive meaning and is socially acceptable to all. \\
Example: "And old customers, there is nothing left for them.. A failed marketing method. Your customer who has been with you for years, discount the value of the invoice."

\textbf{Hate Speech Target:}
We distinguished offensive content as hate speech when targeted at a specific group. We further labeled hateful content into seven categories. Hate speech targeting: {(1) Political ideology, sports, etc. (e.g., "Liberals are destroying our country!"); (2) Origin, ethnicity, or country (e.g., "Go back to your country!"); (3) Religion or sect (e.g., "All Muslims are terrorists"); (4) Social class, profession, etc. (e.g., "Homeless people are lazy"); (5) Physical appearance/disability/rich/poor (e.g., "Fat people are disgusting"); (6) Gender (male/female/misogyny) (e.g., "Women belong in the kitchen") and (7) Other (targeting a specific group) (e.g., "LGBTQ+ people are unnatural").}"

Some text target more than one group. For example, "She looks like a Bangladeshi who works in the grocery store in the neighborhood." 
This tweet bears offensive content. It has targeted two groups: "origin, ethnicity, or country" and "physical appearance/disability". 
While annotating each tweet, we consider all the possible target group. Thus, if multiple annotators annotate one tweet offensive and then they have diverse opinion in target group, we combine all the target group. This introduces the data annotation diversity according to dialectal variation. 

\section{Results and Discussion}

\subsection{Offensiveness Analysis}
Table \ref{tab:offensiveness_analysis} discusses the frequency of offensiveness analysis. The majority of tweets (60.36\%) were identified as hateful or offensive, highlighting a significant presence of negative content in the dataset. Clean or socially acceptable tweets counted for 37.19\%, showing that a substantial portion of the content was non-offensive. There are some tweets that do not fall into the two categories. They belong to different meanings and are very few in number. For instance, a small portion of the tweets are labeled as vulgar (0.018\%), while others contain violent content (0.006\%). Additionally, a few tweets exhibit irony or sarcasm (0.002\%) and humor (0.002\%), highlighting the rarity of these categories.

\begin{table}[h!]
    \centering
    \begin{tabular}{lrr}
        \toprule
        \textbf{Offensiveness} & \textbf{N} & \textbf{\%}  \\ \midrule
        Hateful/Offensive & 6036 & 60.36 \\ 
        Clean/ Socially Acceptable & 3719 & 37.19 \\ 
        Violence & 63 & .006\\ 
        Irony/sarcasm & 26 & .002\\
        Humor & 20 & .002\\ 
        Vulgar & 184 & .018\\ \bottomrule
    \end{tabular}
    \caption{Offensiveness Analysis}
    \label{tab:offensiveness_analysis}
\end{table}

\subsection{Hate Speech Target}
As discussed before, we further classified the offensive content (6036 tweets) into hate speech target group.   
Table \ref{tab:hate_speech_analysis} provides an analysis of the targets of hate speech in tweets, categorized into seven distinct labels. The most common target was political ideology, sports, etc., accounting for 28.76\%. The "Other" category, encompassing a specific group, was almost as prevalent, with 28.51\% of the tweets. Hate speech targeting origin, ethnicity, or country was also notable, representing 26.20\% of the tweets, highlighting significant issues related to racism and xenophobia. Religion or sect was the target in 13.51\% of the tweets, reflecting religious intolerance. Hate speech aimed at social class, profession, etc., appeared in 2.47\% of the tweets. Physical appearance, disability, wealth status, and gender were less frequent targets, with 0.25\% and 0.30\%, respectively. Table \ref{tab:hate_speech_analysis}
shows the hate speech target results.
\begin{table}[h!]
    \centering
    \begin{tabular}{lrr}
       \toprule
        \textbf{Hate Speech Target} & \textbf{N} & \textbf{\%} \\ \midrule
        Political ideology, sports, etc. & 1992 & 28.76 \\ 
        Other & 1975 & 28.51 \\ 
        Origin, ethnicity, or country & 1815 & 26.20 \\ 
        Religion or sect & 936 & 13.51 \\ 
        Social class, profession, etc. & 171 & 2.47 \\ 
        Physical appearance/disability/rich/poor & 17 & 0.25 \\
        Gender (male/female/misogyny) & 21 & 0.30 \\ \bottomrule
    \end{tabular}
    \caption{Hate Speech Target Analysis}
    \label{tab:hate_speech_analysis}
\end{table}

\subsection{Inter Annotator Agreement}
To measure inter annotator agreement, we took a sample of 500 tweets and asked all the annotators to annotate the tweets into two categories: offensive or not (we didn't measure for hate speech target as we combine all possible annotation label to maximize the target group).  
As there were more than two annotators involves in this procedure, we applied Fleiss' Kappa to calculate Inter Annotator Agreement (IAA). 

\begin{equation}
\kappa = \frac{P_o - P_e}{1 - P_e}
\end{equation}

where:

\begin{equation}
P_o = \frac{1}{N \cdot (N - 1)} \sum_{i=1}^{N} \left[ \sum_{j=1}^{N} \left( P_{ij}^2 - \frac{P_i \cdot P_j}{(N - 1)} \right) \right]
\end{equation}

and

\begin{equation}
P_e = \frac{1}{N \cdot (N - 1)} \sum_{i=1}^{N} \left( \sum_{j=1}^{N} \left( \frac{P_i \cdot P_j}{N} \right) \right)
\end{equation}

where \( N \) is the number of raters, \( P_{ij} \) is the proportion of times that the \(i\)-th rater assigned a particular category \(j\), and \( P_i \) and \( P_j \) are the proportions of times categories \(i\) and \(j\) were assigned, respectively.

We achieved an Inter-Annotator Agreement (IAA) score of 0.8143 among the annotators, which indicates a high level of agreement among them. A Fleiss' Kappa value of 0.8143 suggests that the agreement among the annotators is substantial, reflecting a high consistency in their evaluations. 

\subsection{Annotation Evaluation with Transformer Model}
To evaluate the data annotation task, we implemented transformer based model to classify tweet into offensive or not. We also applied transformer model for hate speech target classification. We chose transformer model because it has shown better performance in hate speech classification \cite{alsafari2020hate,Albadi2018are,khezzar2023arhatedetector}. 

\textbf{Model Architecture:} 
Firstly, we prepossessed the text by removing link, username, empty spaces, emojis and weird characters. Then, we applied one hot encoding to each label. We separated the dataset into training:testing:validation as 0.7:0.2:0.1. We used the training dataset to fine tune the pre-trained BERT model. We validate the model outcomes thorugh validation dataset. Finally, we used test dataset to predict the classification layer. The model was set up with 
batch=4, epoch=3, learning rate=2e-5, loss function:sigmoid. We tried with various BERT base model. Initally, we fine-tuned the newest version of AraBERT model which is 'aubmindlab/bert-base-arabertv2' \cite{antoun2020arabert}. Then we fine-tuned with pre-trained CAMelBERT \cite{inoue-etal-2021-interplay} model. We also fine-tuned with XLM-RoBERTa \cite{conneau2019unsupervised}.

\textbf{Model Evaluation:}
Table \ref{tab:performance_metrics} shows the performance metrics for different classifications.
Offensive Vs Clean is binary classification problem. The models are quite good to accurately predict the outcomes. In this cases, AraBERT pre-trained model outperformed with F1-score of 0.7865 and accuracy of 0.7860. CamelBERT also performs well but slightly less effectively than AraBERT.

For hate speech target classification, AraBERTv2 outperforms in all metrics with F1-score 0.6889, Accuracy 0.5985, Precision 0.7172 and Recall 0.6628. It provides a better balance between precision and recall, making it more effective at identifying hate speech targets overall.

\begin{table}[h!]
    \centering
    \small 
    \begin{tabular}{p{1.5cm}p{1cm}p{1cm}p{1cm}p{1cm}}
    \toprule
    \multicolumn{5}{c}{\textbf{Offensive Vs Clean}} \\
    \midrule
    \textbf{Model} & \textbf{F1-Score} & \textbf{Accuracy} & \textbf{Precision} & \textbf{Recall} \\
    \midrule
    AraBERT       & 0.7865 & 0.7860 & 0.7865 & 0.7865 \\
    CamelBERT     & 0.7360 & 0.7355 & 0.7360 & 0.7360 \\
    xlm-roberta   & 0.6280 & 0.6280 & 0.6280 & 0.6280 \\
    \midrule
    \multicolumn{5}{c}{\textbf{Hate Speech Target}} \\
    \midrule
    xlm-roberta   & 0.6607 & 0.5771 & 0.7098 & 0.6181 \\
    AraBERTv2     & 0.6889 & 0.5985 & 0.7172 & 0.6628 \\
    CamelBERT     & 0.6814 & 0.6002 & 0.7234 & 0.6440 \\
    \bottomrule
    \end{tabular}
    \caption{Performance metrics for Offensive vs Clean and Hate Speech Target classification}
    \label{tab:performance_metrics}
\end{table}

\textbf{System Architecture:}
The training was conducted on a Google Colab Server with NVIDIA Tesla T4 GPU with 16GB of VRAM. The experiment was run several times with hyperparameter tuning like epoch, batch size, learning rate, weighted decay and so on. We saved the best weight by comparing the Micro F1-score and predicted the output with the best weight model.

\section{Limitation and Future Scope}
Our hate speech dataset collection is not beyond limitation. 
\begin{itemize}
\item \textbf{Dataset Size:} Considering the size of the dataset, it is limited. While every day people posts a lots in the social media, we need to increase the volume of the dataset. 

\item \textbf{Costly:} Before X (Former Twitter) was free accessible to the research communities. However, it is no more free to researchers and academician. Now everyone has to pay to use the X API to down data which has put limitation on doing research on X data.

\item \textbf{Dialectal variation}: Even though our dataset contains data from all over region, we didn't distuinsh the data based on dialectal region. It reduces the model performance while predicting the outcomes. 

\end{itemize}

\section{Data and Code Availability}
The annotated dataset are available on \url{10.5281/zenodo.14669917}. The dataset is released under the CC BY 4.0 license, allowing free use, distribution, and adaptation with proper attribution. The code is available in \url{https://github.com/rafiulbiswas/hatespeech-detection}.

\section{Ethical Consideration}
We obtained an institutional IRB exemption for this project. This study complies with Twitter API rules and protects user privacy by anonymizing personally identifiable information. Only public tweets were collected, and the dataset may not represent all Twitter users. Data was handled securely and used only for this study. The study adheres to ethical guidelines for social media research.

\bibliography{aaai25}

\end{document}